\documentclass[nohyperref]{article}

\usepackage{microtype}
\usepackage{graphicx}
\usepackage{subfigure}
\usepackage{xspace}
\usepackage{booktabs} 
\usepackage{hyperref}

\newcommand{\ours}{\textsc{CAMELoT}\xspace}

\usepackage[accepted]{icml2022}

\usepackage{amsmath}
\usepackage{amssymb}
\usepackage{mathtools}
\usepackage{amsthm}
\usepackage{multirow}

\usepackage[capitalize,noabbrev]{cleveref}
\crefname{section}{Section}{Sections}
\crefname{subsection}{Section}{Sections}
\crefname{subsubsection}{Section}{Sections}
\crefname{figure}{Figure}{Figures}
\crefname{table}{Table}{Tables}
\crefname{subfigure}{Figure}{Figures}

\theoremstyle{plain}

\theoremstyle{definition}

\theoremstyle{remark}

\usepackage{newfloat}
\usepackage{listings}
\usepackage{booktabs}
\usepackage{graphicx}
\usepackage[textsize=tiny]{todonotes}

\icmltitlerunning{CAMELoT: Towards Large Language Models with Training-Free Consolidated Associative Memory}

\begin{document}

\twocolumn[
\icmltitle{\ours: Towards Large Language Models \\with Training-Free Consolidated Associative Memory}



\icmlsetsymbol{equal}{*}

\begin{icmlauthorlist}
\icmlauthor{Zexue He}{ucsd}
\icmlauthor{Leonid Karlinsky}{ibm}
\icmlauthor{Donghyun Kim}{korea}
\icmlauthor{Julian McAuley}{ucsd}
\icmlauthor{Dmitry Krotov}{ibm}
\icmlauthor{Rogerio Feris}{ibm}
\end{icmlauthorlist}

\icmlaffiliation{ucsd}{Department of XXX, University of YYY, Location, Country}
\icmlaffiliation{ibm}{School of ZZZ, Institute of WWW, Location, Country}
\icmlaffiliation{korea}{Company Name, Location, Country}

\icmlcorrespondingauthor{Firstname1 Lastname1}{first1.last1@xxx.edu}
\icmlcorrespondingauthor{Firstname2 Lastname2}{first2.last2@www.uk}

\icmlkeywords{Machine Learning, ICML}

\vskip 0.2in
\center{UC San Diego \textsuperscript{1} \hspace{0.2in}  MIT-IBM Watson AI Lab, IBM Research \textsuperscript{2} \hspace{0.2in}  Korea University \textsuperscript{3}}
\vskip 0.2in
]




\begin{abstract}


Large Language Models (LLMs) struggle to handle long input sequences due to high memory and runtime costs. Memory-augmented models have emerged as a promising solution to this problem, but current methods are hindered by limited memory capacity  and require costly re-training to integrate with a new LLM.  In this work, we introduce an associative memory module which can be coupled to any pre-trained (frozen) attention-based LLM without re-training, enabling it to handle arbitrarily long input sequences. Unlike previous methods, our associative memory module consolidates representations of individual tokens into a non-parametric distribution model, dynamically managed by properly balancing the novelty and recency of the incoming data. By retrieving information from this consolidated associative memory, the base LLM can achieve significant (up to 29.7\% on Arxiv) perplexity reduction in
long-context modeling compared to other baselines evaluated on standard benchmarks. This architecture, which we call CAMELoT (\textbf{C}onsolidated \textbf{A}ssociative \textbf{M}emory \textbf{E}nhanced \textbf{Lo}ng \textbf{T}ransformer), demonstrates superior performance even with a tiny context window of 128 tokens, and also enables improved in-context learning with a much larger set of demonstrations.

 
\end{abstract}

\section{Introduction}
\begin{figure}[t]
\begin{center}
\includegraphics[width = \linewidth]{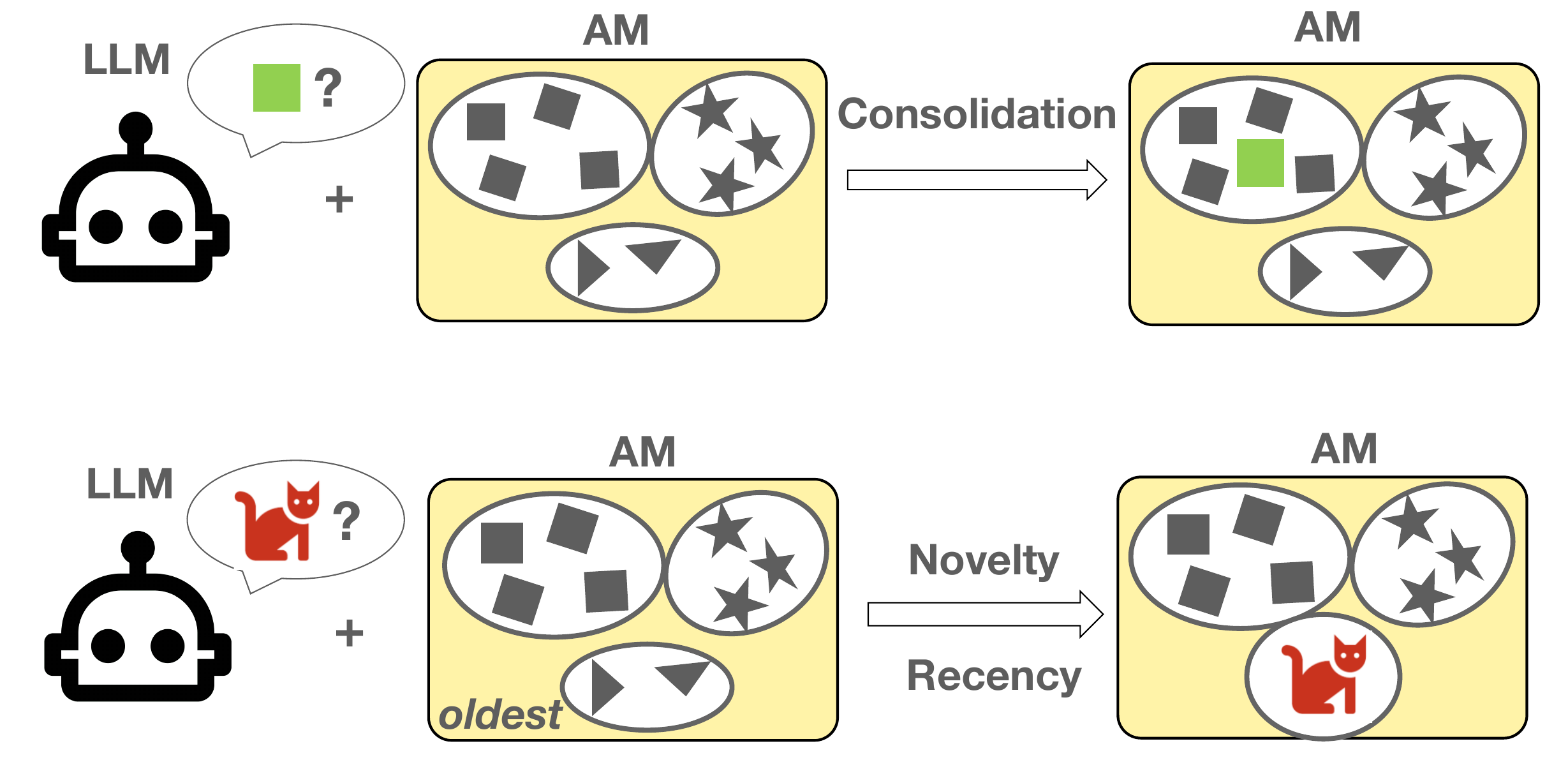}
\end{center}
\caption{\textbf{C}onsolidated \textbf{A}ssociative \textbf{M}emory \textbf{E}nhanced \textbf{Lo}ng \textbf{T}ransformer 
 (\ours). Top: Consolidation of representations in the associative memory (AM) -- related concepts are grouped together and averaged. Bottom: Recency-dependent incorporation of novel concepts -- when a new concept is introduced with no close matches,  the oldest slot (since its last update) is replaced with the new concept.} \label{fig:teaser}
\end{figure}

\begin{figure}[t]
\begin{center}
\includegraphics[width = \linewidth]{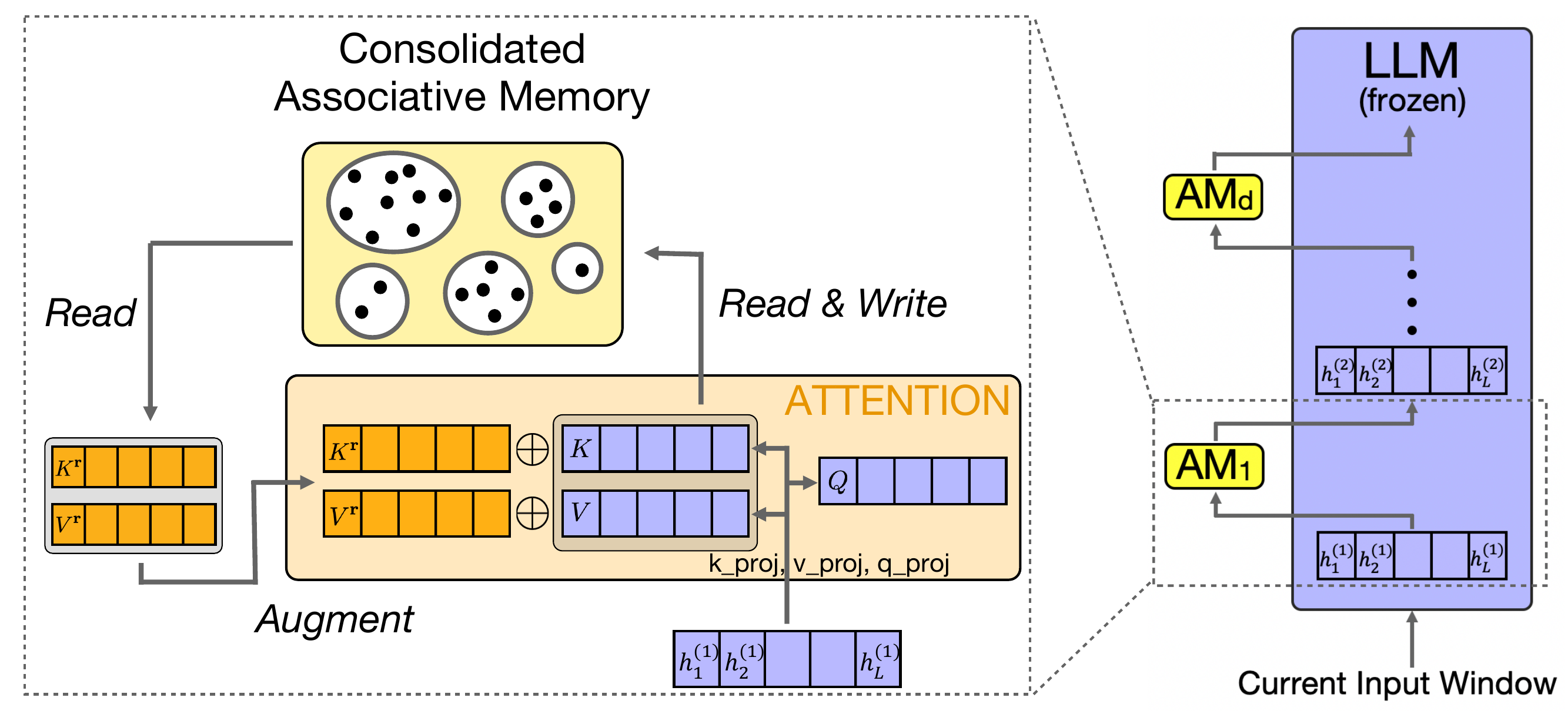}
\end{center}
\caption{The general pipeline of our method. Every layer of the backbone LLM is augmented with an AM module (we draw AM in the first attention layer here, just as an example). Keys and values are calculated for every token, keys are used to search for relevant memorized tokens in the memory bank and return them (Read). The retrieved memory keys and values are prepended to the original token keys and values as prefixes. Finally, the attention operation is applied on the concatenation of the retrieved and native keys and values (Augment). After retrieval, the memory state is modified according to the Write operation, see \autoref{fig:read&write}.} 
\vspace{-0.5em}
\label{fig:pipeline}
\end{figure}

Throughout life, humans are exposed to a myriad of events perceived through their sensory systems. Over time, these individual events are processed and consolidated to form memories, which exemplify groups of related events and form the basis for future actions. Through consolidation, individual data instances turn into memorized knowledge, which represents general aspects of individual events but discards inessential details \cite{sara2000retrieval}.  

Associative Memory (AM) is an important class of memory systems. The core computation of AM is to link (associate) a query with representations stored in the memory \cite{willshaw1969non,hopfield1982neural}. Specifically, for any given query AM should pick the {\it consolidated} memory slot with the representation that matches the query in the best possible way. These stored representations should concisely summarize past experiences and provide valuable cues for future actions. Recently there has been a significant interest in modern reincarnations of associative memory networks \cite{krotov2016dense,ramsauer2021hopfield}. There is a large body of literature on memory consolidation in neural networks \cite{dudai2004neurobiology}, as well as a variety of local learning rules, which are computationally cheaper than the end-to-end backpropagation training \cite{tyulmankov2021biological}. 

Concurrently, large language models (LLMs) have become very important for many practical applications such as chatbots, text generation \cite{radford2019language}, theorem proving \cite{polu2020generative}, quantitative reasoning \cite{lewkowycz2022solving}, and question answering \cite{chung2022scaling}, etc. One crucial parameter for LLMs is the allowed input context length $L$. Supporting longer context makes it possible to incorporate richer factual information and work with long documents or large code repositories, which has been observed to increase LLM performance in cases when this longer context is available at inference time \cite{press2022train}. Unfortunately, the conventional transformer attention mechanism scales quadratically ($L^2$) with an increasing number of tokens, which makes increasing the context length 
computationally
hard.


Besides, training state-of-the-art LLMs is an expensive endeavour that usually requires large compute resources, big engineering teams, etc., for which reason is generally available only to large organizations. Open source LLMs, although work well in practice, often have limited context windows, facing challenges with NLP tasks involving long documents.
These constraints raise a question: \textit{is it possible to develop a plug-and-play module that can be coupled to a pre-trained (frozen) LLM enabling it to handle (unlimited) long context far beyond $L$}? Importantly, this module should be less computationally cost and should not require {\it any} retraining or fine-tuning of the backbone LLM. 


Our work tackles this question and is inspired by ideas from Associative Memory(AM). We propose a  plug-and-play module which consolidates representations of individual tokens in AM in a way that depends on the novelty and recency of the concepts (as shown in \cref{fig:teaser}). Our module can be coupled to {\it any} pre-trained (without such a memory module) attention-based LLM. The function of the module is to consolidate information about the prior context far beyond the current context window (of length $L$) while approximating the attention computation (for the next token prediction) on the full (unlimited) context. The information is written into the AM module using a local writing rule, which is computationally cheap, 
and dynamically keeps track of recency and novelty of the incoming information. 

The consolidated context is modeled as a collection of non-parametric distributions, one per key-space of every LLM layer. The modes of those distributions are consolidated dynamically as the context window sweeps over the long overall context, new modes are created according to their detected novelty, while outdated modes are removed. The long context attention is then simply approximated via retrieving modes closest to the current context hidden states and adding them in the form of a key-value cache.
%
Our method does not require any re-training, fine-tuning, or learning adaptors between the base LLM and the AM module. We demonstrate that this simple strategy leads to  significantly strong results on the language modeling tasks and the in-context learning task. For instance, when coupled to a pre-trained LLaMA model our memory-enhanced network results in significant (up to 29.7\% on Arxiv) perplexity reduction in long-context modeling compared to the base LLM.

\section{Related Work} 

{\bf Long-range self-attention.} Many efficient self-attention techniques have been proposed for long context modeling in transformer models, including low-rank factorization \cite{linformer}, local attention \cite{ramachandran2019stand}, dilated attention \cite{longnet}, sparsity \cite{longformer,bigbird,reformer}, and hardware-aware attention mechanisms such as FlashAttention \cite{flashattention1, flashattention2}. Despite notable progress, these methods are unable to handle unbounded context window sizes and struggle to retrieve information in the middle of the input \cite{liu2023lost}. They can be used in tandem with our proposed approach for longer context modeling.

{\bf Memory-augmented LLMs.} Memory-augmented language models have emerged as a promising way to model extended context window sizes \cite{memgpt,transformerxl,memorizing,longllama,memorynetworks}. In particular, \citet{memorizing} show that a kNN lookup into a memory cache bank containing (key, value) pairs of past inputs can improve language modeling. \citet{longllama} further improved this approach using contrastive learning. In the same vein,  \citet{longmem} addressed the memory staleness limitation of these works by training a side network model, while keeping the LLM frozen. Unlike these methods, our approach relies on consolidated representations of past tokens which are dynamically updated, enabling the context window to be arbitrarily large, without being limited by the number of memory slots. Moreover, different from these approaches, our method is {\em training-free} (memory updates occur solely at runtime), making it easier to integrate our memory module into any existing LLM architecture.

{\bf Context compression.} Prompt compression techniques \cite{ICAE,gist,autocompressor} have been recently explored for extending the context length in transformer models. These methods operate at the input level, while our method consolidates the internal representations of the model based on a local associative memory update rule. \citet{compressivetrans} proposed the Compressive transformer, which compresses past activations of the model for long-range sequence modeling. In contrast, our proposed approach does not require training or additional losses like attention-reconstruction. In addition, we offer a novel way to effectively update our associative memory representations, balancing information about novelty and temporal proximity. 

{\bf Memory networks.} There is a large body of literature on memory models, e.g.~memory networks \cite{memorynetworks}, sparse distributed memory \cite{kanerva1988sparse}, and various forms of associative memory \cite{kohonen2012associative}. Neuroscience-inspired memory models have also been used for language model augmentation \cite{park2023memoria}. Memory augmentation has also been used in reinforcement learning settings \cite{graves2016hybrid}, and successfully coupled to recurrent neural networks \cite{graves2014neural}. To the best of our knowledge, none of these works enable consolidated memory augmentation of LLMs without requiring additional training, as in our approach.

\section{Associative Memory (AM)-enabled LLM}
\begin{figure*}[t]
\begin{center}
\includegraphics[width = 1.0\linewidth]{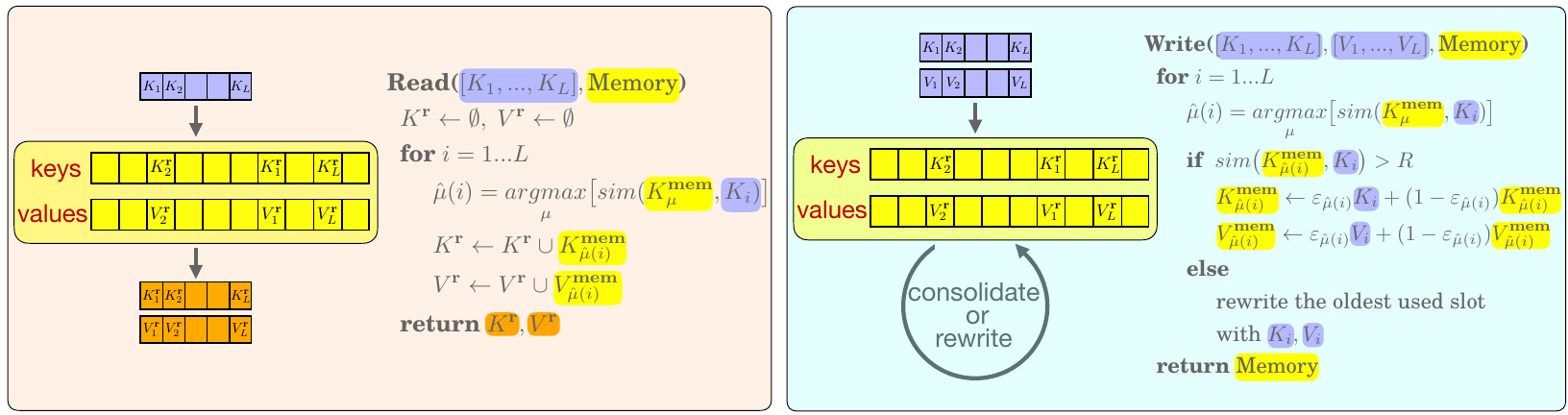}
\end{center}
\vspace{-4mm}
\caption{Every AM module performs read and write operations. The read operation retrieves memorized tokens most similar to the native keys. The write operation updates the state of the memory by performing consolidation, which depends on novelty and recency.} \label{fig:read&write}
\end{figure*}
For long document modelling tasks it is desirable to have an architecture capable of efficient usage of information that appeared in past contexts. Our proposed method is built on three desiderata. First, redundant information from the past should be compressed and stored in the AM block while reducing repetitions ({\bf consolidation}). When the same concept appears in the past context multiple times, it is wasteful to store each individual instance of that concept in a separate memory slot; instead, all those instances should be consolidated and stored only once. Second, novel concepts not
encountered by the model in the past must be detected and stored in a new memory slot at their first encounter ({\bf novelty}).
These novel memory slots can be subsequently consolidated with the possible future occurrences of related concepts. Third, in situations when the topic shifts, the model should be able to discard outdated memory slots that are no longer useful, if that is required for the incorporation of additional novel concepts encountered following the topic shift ({\bf recency}).     

To achieve these desiderata, we design \ours, a \textbf{C}onsolidated \textbf{A}ssociative \textbf{M}emory \textbf{E}nhanced \textbf{Lo}ng \textbf{T}ransformer, consisting of a base language model and a memory module (overall architecture shown in \autoref{fig:pipeline}). The memory module is equipped with a \textbf{Read} and \textbf{Write} operation, supporting information retrieval from the memory bank and the update to the memory bank. With the retrieved information, the  current context window of LLM is memory-enhanced via the \textbf{Augment} operation. 
These three desiderata are the foundation of \ours. 
Our method is agnostic to the specific choice of many popular transformer architectures, in the sense that any attention-based LLM can be enhanced with the AM in \ours.  

\subsection{Read Operation}
When a context window of length $L$ is processed through the LLM, keys and values from every layer (more generally can be an arbitrary subset of layers) are passed to the corresponding AM module (one per memory-augmented layer). AM in each layer consists of $M$ memory slots, enumerated by the index $\mu = 1,...,M$. Each slot contains two vector variables: memory keys $K^\mathbf{mem}_\mu$ and memory values $V^\mathbf{mem}_\mu$, and two integer scalar variables: counts $c_\mu$ (number of consolidated instances), and age $\tau_\mu$ (how old the current slot is since its last update). 

When a set of keys $K_i$ and values $V_i$ (index $i=1,...,L$ enumerates individual tokens from the current context window) is passed to the AM module to retrieve relevant information, a \textit{search function} identifies the memory slots with the strongest association (highest similarity) between the input token key $K_i$ and AM's memory slot keys $\{K^\mathbf{mem}_\mu\}$: 
\begin{equation}
    \hat{\mu}(i) = \underset{\mu}{argmax} \big[sim(K^\mathbf{mem}_\mu, K_i)\big]
\end{equation}
The keys and their corresponding values of these $L$ strongest-associated memories ($K^\mathbf{r}$ and $V^\mathbf{r}$) are returned for the current $L$ native tokens and passed back to the LLM in the form of the key-value cache. See details in \cref{fig:read&write}.

\subsection{Augment Operation}
The list of retrieved key-value caches ($K^\mathbf{r}$ and $V^\mathbf{r}$) are passed back to the base LLM and  used as the prefix context in each respective memory-augmented layer. They are prepended to the LLM keys and values of current input tokens. Then causal attention is performed on the concatenated list, which after the augmentation contains $2L$ keys and values (the length of current native context + the length of retrieved memories) and $L$ queries (current context only), resulting in the augmented transformer attention output $[a_1, \cdots, a_L]$. The attention output results in augmented hidden states $[h_1, \cdots, h_L]$ which are the input to the next layer, as shown in the following equations and \autoref{fig:read&write}:
\begin{align}
    [a_1,  \cdots, a_L] &= Attn(Q, K', V') \\
    Q &=[Q_1, Q_2, \cdots, Q_L] \\
    K'&=  K^\mathbf{r}	\oplus [K_1, \cdot, K_L],\\
    V' &=  V^\mathbf{r} 	\oplus [V_1, \cdots, V_L] 
    \label{eq:attention}
\end{align}

\subsection{Write Operation}
The state of AM is updated by the current context window according to the \textbf{Write} operation comprised of two parts explained next (see \autoref{fig:read&write} for an illustration).  

\textbf{Consolidation.} If the similarity between the current context token key and the strongest-associated memorized key is large ($>R$, $R$ is a hyper-parameter), the concept described by that token is declared familiar and, for this reason, its key and value are consolidated with the key and value stored in that memory slot. Specifically, memory slots are updated according to:    
\begin{align}
K^{\mathbf{mem}}_{\hat{\mu}(i)} &\leftarrow \frac{K_i + c_{\hat{\mu}(i)}K^{\mathbf{mem}}_{\hat{\mu}(i)}}{c_{\hat{\mu}(i)}+1}  \\
V^{\mathbf{mem}}_{\hat{\mu}(i)} &\leftarrow \frac{V_i + c_{\hat{\mu}(i)}V^{\mathbf{mem}}_{\hat{\mu}(i)}}{c_{\hat{\mu}(i)}+1} \\
c_{\hat{\mu}(i)} &\leftarrow c_{\hat{\mu}(i)}+1
\end{align}
where $c_\mu$ tracks the number of instances consolidated in slot $\mu$. Thus, the consolidated representations stored in each slot $\mu$ are always arithmetic averages of individual instances that went into that slot. By introducing an update rate $\varepsilon_\mu = 1/(c_\mu +1)$, these expressions can be rewritten as incremental modifications to the existing representations stored in the AM.


\textbf{Novelty and Recency.} If the similarity with the closest memorized key is weak ($<R$), the concept is declared novel. In this case, the oldest unused memory slot (the one with maximal age $\tau_{\mu}$) is replaced with $K_i$, $V_i$, and its age is set to $0$. 
After each slot $\hat{\mu}(i)$ update and its age statistic $\tau_{\hat{\mu}(i)}$ is set to $0$,  the ages of all slots that had no matching current context hidden state are incremented by $1$.



\subsection{Probabilistic interpretation}
Alternatively, the keys and values in AM slots can be viewed as modes of a non-parametric Gaussian mixture distribution estimation approximating the key-value manifold of the past context windows. This mixture accepts new key-value points from the current context via a diagonal kernel of width $\hat{R}$ (distance measure corresponding to similarity $R$). The means (centers) of the modes of the mixture are updated according to the above online average rules while maintaining the needed sufficient statistics (counts) for computing the averages in further updates. Retrieving nearest distribution modes to the current context hidden states effectively  approximates the full (long) context attention, at least within the radius $\hat{R}$ from the retrieved mode centers. For the tokens whose keys and values are beyond radius $\hat{R}$ of their closest mode, new modes are created online, while the oldest modes are evicted, maintaining the recency of our distribution estimation and its correspondence with the evolving context.

\section{Experiments}
\paragraph{Evaluation Scenarios.}
We evaluate \ours on two long text modeling scenarios:  \textit{causal language modeling (CLM) tasks } and \textit{few-shot in-context learning (ICL) tasks}. 

In CLM tasks, we follow the data preprocessing method proposed by \citet{transformerxl} where lengthy documents are segmented into sequential, non-overlapping windows and the LLM processes each window one by one. During this process, we first use the key and value representations of each token to read from the AM and retrieve the relative information, then augment the causal language model by taking the returned memory as the past caches. Subsequently, the keys and values of the current context input are integrated into the AM via the Write function. We measure the perplexity for tokens in each window and calculate their average across the entire long context in the end. 

In the few-shot ICL tasks, we similarly split the few-shot examples in a long prompt into non-overlapping inputs and process each one individually. We sequentially extract the keys and values of each input window from the language model, and then add them into the AM through the Write function. Note that we do not perform retrieval for input windows that are part of the context examples. Afterwards,  after getting the keys and values of the test question in the current long prompt from the language model, we conduct the Read operation for each generated token and the retrieved memory caches are used to augment the answer generation. 


\paragraph{Details.} We take the officially released LLaMa2-7b from Huggingface Library as the base model in \ours. For the two experiments, we set the size of memory banks to be 10K slots. We put memory banks into a single NVIDIA-A100 GPU for fast parallel computation. We also notice that one can use FAISS \cite{johnson2017billionscale} approximate search, which is a simple extension of our framework. We use cosine similarity in \ours and the similarity threshold $R$ in novelty detection is set to be 0.93. Unless specified otherwise, our experimental results are reported for \ours with 10k memory slots (more experiments on memory size are detailed in \cref{Appendix:more_memory_size}).  We preprocess and batch the evaluation corpus following the standard implementation of Transformer-XL \cite{transformerxl}.  For more details, please refer to ~\cref{Appendix: hyper}. 


\subsection{AM-augmented Long Language Modeling}

\begin{table*}[ht]
\centering

\begin{tabular}{@{}ccccc|ccc@{}}
\toprule
                                & \multicolumn{4}{c|}{{PG-19 and Arxiv}} & \multicolumn{3}{c}{{wikitext-103}} \\ \cmidrule(lr){1-5}\cmidrule(lr){6-8}
 &
  {Input Length} &
  {Retrieved Mem.} &
  {PG-19} &
  {Arxiv} &
  {Input Length} &
  {Retrieved Mem.} &
  {Wikitext-103} \\ \cmidrule(lr){1-5}\cmidrule(lr){6-8}
\multirow{4}{*}{{LLaMa2-7B}}      & 512       & None   & 9.54   & 5.99  & 512       & None   & 16.0   \\
                                & 1024      & None   & 8.33   & 4.98  & 1024      & None   & 14.80  \\
                                & 2048      & None   & 7.88   & 4.35  & 2048      & None   & 14.46  \\
 & Avg       & - &
  8.58 &
  5.12 & Avg       & - &
  15.09 \\ \midrule
\multirow{4}{*}{Transformer-XL} & 512       & 512    & 8.44   & 4.15  & 256       & 256    & 15.02  \\
                                & 1024      & 1024   & 8.27   & 3.81  & 512       & 512    & 14.21  \\
                                & 2048      & 2048   & 7.86   & 3.65  & 1024      & 1024   & 14.2   \\
                                & Avg       & -      & 8.19   & 3.87  & Avg       & -      & 14.48  \\\cmidrule(lr){1-5}\cmidrule(lr){6-8}
\multirow{4}{*}{\begin{tabular}[c]{@{}c@{}}Memorizing \\ Transformers\end{tabular}} &
  512 &
  512 &
  8.12 &
  3.82 &
  256 &
  256 &
  14.18 \\
                                & 1024      & 1024   & 7.4    & 3.63  & 512       & 512    & 14.07  \\
                                & 2048      & 2048   & 7.34   & 3.62  & 1024      & 1024   & 14.39  \\
                                & Avg       & -      & 7.62   & 3.69  & Avg   & -      & 14.21  \\ \cmidrule(lr){1-5}\cmidrule(lr){6-8}
\multirow{4}{*}{CAMELoT}        & 512       & 512    & 7.24   & 3.61  & 256       & 256    & 14.06  \\
                                & 1024      & 1024   & 7.14   & \textbf{3.60}  & 512       & 512    & \textbf{14.00}  \\
                                & 2048      & 2048   &\textbf{ 7.10}   & \textbf{3.60 } & 1024      & 1024   & 14.34  \\
                                & Avg   & -      & 7.16   & 3.60  & Avg       & -      & 14.13 \\ \bottomrule
\end{tabular}
\caption{Language Modeling Perplexity on wikitext-103, Arxiv, and Pg-19. For wikitext-103, we notice the maximum length of its documents is smaller than 2k. Therefore, we report results of models whose effective input length $\leq$ 2048 (i.e., input length $\leq$ 2048 for non-augmented model; and input length $\leq$ 1024 for memory-augmented models). \textbf{Bold}: Best perplexity on each dataset. Avg: Average.} 
\label{tab:llm-w-baseline}
\end{table*}

\subsubsection{Setup}
We evaluate the long context language modeling capabilities of \ours using three key datasets: 

\textbf{Wiki-103} \cite{merity2016pointer}\footnote{https://blog.salesforceairesearch.com/the-wikitext-long-\\term-dependency-language-modeling-dataset/}, which comprises articles from Wikipedia covering various topics with  good language quality;\newline
\textbf{Arxiv} \cite{pile}\footnote{Taken from the Pile: https://pile.eleuther.ai/}, a collection of academic papers primarily in the fields of Mathematics, Computer Science, and Physics.  This dataset is recognized for its high-quality text and mathematical content, making it a challenging benchmark for long-context language modeling; 
\newline 
\textbf{PG-19} \cite{raecompressive2019}\footnote{https://github.com/google-deepmind/pg19} which includes full-length books offering a standard benchmark widely used in long-range natural language modeling \cite{memorizing,longmem,longllama}. 


We take the test split of each dataset and report its language modeling perplexity.

In our experiment, we benchmark \ours against two notable memory-augmented transformers that have demonstrated effectiveness in long language modeling tasks:

\textbf{Transformer-XL} \cite{transformerxl}: This model uses a finetuning-based approach, storing a fixed length of previous input in a cache to enhance the current input. Notably, it does not employ similarity-based retrieval. \newline
\textbf{Memorizing Transformer} \cite{memorizing}: this model saves past caches in a circular manner. Thus older caches are replaced by newer ones as the memory bank fills up (no consolidation occurs) and similar caches are retrieved for input augmentation. The official implementation relied on fine-tuning.

For a fair comparison, in \ours and the baselines experiments, we used the same LLaMa2-7B backbone (original baselines used weaker backbones, such as GPT2), and did not use fine-tuning.

\subsubsection{Performance Comparision}
\cref{tab:llm-w-baseline} compares \ours with the baseline models. While memory-augmented methods generally improve upon the base model on test perplexity, our analysis uncovers the following disparities in their effectiveness. Transformer-XL shows the least improvement, hindered by the lack of relevance assessment during memory augmentation. The Memorizing Transformer, with its capability to selectively retrieve relevant information from the past, outperforms Transformer-XL. However, it lacks memory consolidation, meaning it can only hold a finite cache before older memories are overwritten, limiting its long-term utility.

By not only selecting relevant past information but also employing a novel memory consolidation process, \ours significantly enhances model performance 
($16.6\%$ on PG-19, and $29.7\%$ on Arxiv, and $6.36\%$ on wikitext-103, relative the base model on average), surpassing other memory-augmented methods. Remarkably, \ours achieves superior performance at shorter input lengths, demonstrating its handling of long-range dependency regardless of input size. We provide further discussion of this phenomenon in \cref{sec:discussion_shorter_input}.

These findings underscore \ours's  ability to consolidate and retrieve information from its associative memory, balancing recency and novelty effectively to support modeling long-range dependencies.

\begin{table}[t]
\centering
\begin{tabular}{@{}ccc@{}}
\toprule
                                  & {Prompt Length} & {Accuracy} \\ \midrule
{LLaMa2-7b}                & 13 (0-shot)                 & 0.15              \\ \midrule
\multirow{4}{*}{{CAMELoT}} & 2k                     & 0.21              \\
                                  & 4k                     & 0.25              \\
                                  & 10k                    & 0.42              \\
                                  & 20k                    & 0.44              \\ \bottomrule
\end{tabular}
\caption{Test accuracy on TREC-50 few-shot ICL with various numbers of in-context examples in the prompt (thus various prompt lengths).}
\vspace{-2em}
\label{tab:ICL_performance}
\end{table}


\subsection{In-context Learning Enhanced by AM}
Large language models (LLMs) have demonstrated impressive in-context learning capabilities, offering effective solutions across a range of tasks through few-shot prompting \cite{brown2020language}. We observe that LLM's in-context learning performance improves with the addition of more examples in the prompts \cite{longllama,he2023medeval}. However, this improvement is naturally limited by the maximum input length the models can handle. In this section, we explore the potential of \ours to enhance in-context learning by storing crucial knowledge from in-context examples into the AM module. Using a few-shot question-answering task as our test case, we investigate whether \ours can outperform traditional LLMs by effectively leveraging relevant information in the memory.

\subsubsection{Setup}

Our experiment uses the Text REtrieval Conference (TREC) Question Classification benchmark, which features sentences categorized into 50 labels and has an average sentence length of 10 words. Following \citet{hao2022structured}, we construct each test prompt in the following way: for a given test question, we randomly sample $P$ sentence-label pairs from the training set to form the $P$-shot context. This context is then concatenated with the test question at the end, creating a long test prompt. By adjusting the value of $P$, we can generate test prompts of varying lengths, with detailed examples provided in  \cref{Appendix:ICL_examples}.

For each long test prompt created, \ours starts by storing the context examples in its memory bank. It then proceeds to generate an answer for the test question by retrieving the most relevant information from the updated memory bank. Following \citet{hao2022structured}, \ours selects the option that yields the lowest perplexity among all options and uses it as the prediction to calculate accuracy.


\subsubsection{Result}

\cref{tab:ICL_performance} shows the accuracy of \ours with 10k memory slots on the TREC test set. In this experiment, we observe that adding more examples to the prompt and modeling it with \ours memory module significantly enhances the QA performance. With the memory slots nearly filled to capacity with examples, a large improvement is observed when comparing \ours with 10k slots to its version with 4k slots (another 17\% accuracy added). We also notice the gains diminish when comparing the 20k slots version to the 10k slots one. This diminishing gain is understandable as adding more examples to the memory leads to more instances of conflict, potentially diminishing the marginal benefit of further additions. 


\section{Discussion}
\begin{figure}[th]
	\centering
	\includegraphics[width=0.9\linewidth]{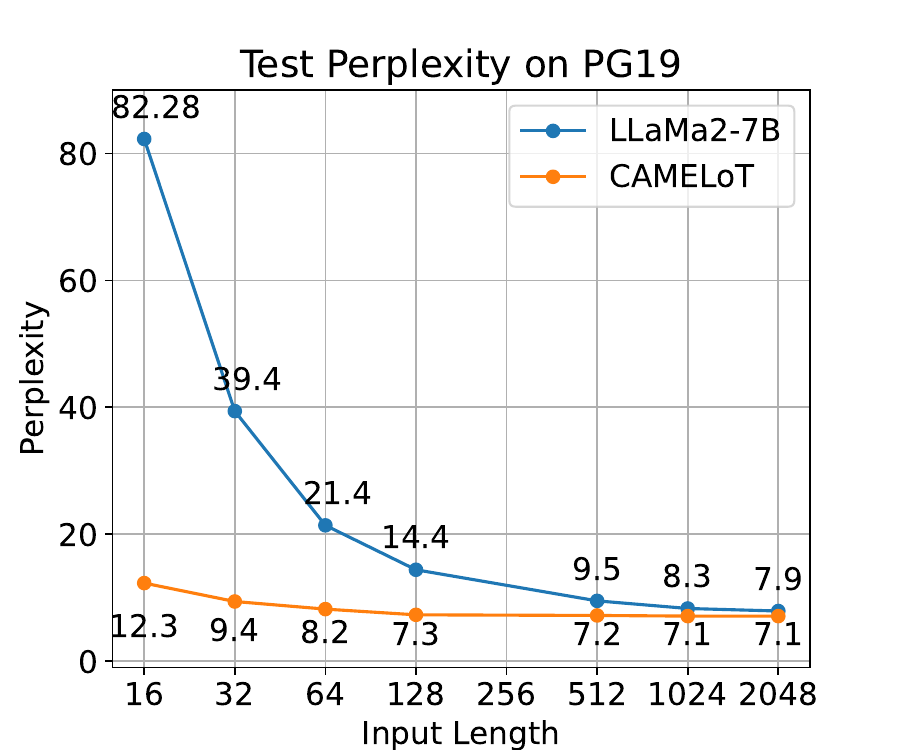}
	\caption{Test perplexity on PG19 with different input lengths.}
	\label{fig:position}
 \label{fig:efficiency}
\end{figure}

\subsection{Shorter Inputs, Better Performance}

\label{sec:discussion_shorter_input}

This section analyzes \ours's performance with different input lengths on the PG-19 test set, using 10k memory slots. Results are shown in \cref{fig:efficiency}.

Unlike models without memory augmentation, \ours demonstrates a relatively consistent performance across different input lengths. This stability can be attributed to the integration of additional knowledge in the AM saved from previous inputs. As \ours accumulates past information, its visible context range extends beyond the current input, allowing an effective modeling of long-range dependencies irrespective of the length of the current input. In contrast, the model lacking memory augmentation relies solely on the local context of the current input, leading to performance fluctuations based on input length.

\ours maintains its effectiveness even with tiny input lengths (e.g., 128), reducing the demand on hardware resources such as large GPUs. This enables transformers to operate attentions with shorter inputs but without compromising the quality of language modeling. Such an advantage lowers the barriers for deploying large language models in environments where  computational budget is limited. 

\subsection{Ablation Analysis} 
\begin{table}[t]
\centering
\begin{tabular}{@{}lc@{}}
\toprule
\multicolumn{1}{c}{\textbf{Models}} & \textbf{PPL}      
\\ \midrule
LLaMa2-7B                  & 7.30             
\\ \cmidrule(lr){1-2}
\ours                       & \textbf{6.85}             
\\ \cmidrule(lr){1-2}
\ours w/o Read              & $>20$ 
\\
\ours w/o Recency             & 9.25            
\\
\ours w/o Novelty            & 7.23            
\\
\ours w/o Consolidation       & 7.00            
\\ \bottomrule
\end{tabular}
\caption{Ablation Study on PG-19-sampled. We report the relative performance lost in perplexity over the full \ours.}
\vspace{-0.5em}
\label{tab:ablation}
\end{table}

\begin{table}[ht]
\centering
\resizebox{1\linewidth}{!}{%
\begin{tabular}{@{}ccccc@{}}
\toprule
\multirow{2}{*}{} & \multicolumn{4}{c}{\textbf{Frequency}}                                              \\ \cmidrule(l){2-5} 
                  & \textgreater{}10K        & 1K - 10K & 100 - 1K & \textless{}100            \\ \midrule
LLaMa2-7B         & \multicolumn{1}{c}{2.75} & 5.08     & 9.77     & \multicolumn{1}{c}{25.96} \\
CAMELoT           & 2.13              & 4.11     & 7.54     & 19.4                     \\ \midrule

\begin{tabular}[c]{@{}c@{}}Relative Gain\\ Over LLaMa2\end{tabular} & 22.1\%            & 19.3\%   & 22.8\%   & 25.3\% \\ \bottomrule
\end{tabular}
}
\caption{Test perplexity broken down by word frequency buckets. }
\vspace{-0.5em}
\label{tab:infrequent_ppl}
\end{table}

To assess the impact of each component within \ours, we define the following ablation variants:

\textbf{\ours w/o Retrieval:} Instead of retrieving the closest matching memory concept for each token in the current input, a random memory concept is returned.\newline
\textbf{\ours w/o Recency:} 
If a token's mode has no close match in memory, it randomly replaces a memory slot rather than the outdated one, ignoring recency.
\newline
\textbf{\ours w/o Novelty.} 
Tokens are consolidated into their closet slot, regardless of if they are from novel modes. R=-1 in cosine similarity retrieval.
\newline
\textbf{\ours w/o Consolidating.} 
 Memory gets updated by token representations based on temporal recency, without consolidating, setting R=+1.

We evaluate on PG-19 Sampled dataset, a subset of PG-19 comprising 20\% of the books in test set. We report test perplexity for each variant with a context length of 2048.


\begin{table*}[ht]
\centering
\begin{tabular}{@{}lll@{}}
\toprule

\multicolumn{1}{c}{\textbf{Slot97: Pronouns}}                                                                                                                                                                                                    & \multicolumn{1}{c}{\textbf{Slot110: Prefix}}                                                                                                                                                                                                & \multicolumn{1}{c}{\textbf{Slot103: Suffix}}                                                                                                                                                                                                                    \\ \midrule
\begin{tabular}[c]{@{}l@{}}this, he, she, her, I, our,  were\\ had, They, their, they, was\\ that, is, are, those, there, these\end{tabular}                                                                                               & \begin{tabular}[c]{@{}l@{}}pre, Re, alt, al, be, bel, del, comp, Al, \\ per,ple, ab, dis, no, non, de, un, im, \\ bl, bri, Ch, Eng, com, fl, Fr, sal, gen, \\ str, et, es\end{tabular}                                           & \begin{tabular}[c]{@{}l@{}}atives, ful, ate, ere, ish, ible, ily,\\ ry, ly, ling, ent, ence, er, ine,  ina, \\ ier, age, ations, ation, ood, inity, \\ itute\end{tabular}                                                                                 \\ \bottomrule\toprule
\multicolumn{1}{c}{\textbf{Slot60: States to Civilization}}                                                                                                                                                                                 & \multicolumn{1}{c}{\textbf{Slot394: Masculine to Feminine}}                                                                                                                                                                                 & \multicolumn{1}{c}{\textbf{Slot7275: Num. to  Adv.}}                                                                                                                                                                                                        \\ \midrule
\begin{tabular}[c]{@{}l@{}}\colorbox{pink}{Minn, Miss, states, Tennessee, PA}, \\ \colorbox{pink}{Minnesota, Lincoln, Pitts, Kingdom,} \\ \colorbox{pink}{Phil}, Si, prep, \colorbox{pink}{DEL, Eng, Montreal}, \\ \colorbox{pink}{British, Franklin, Hill,  }\colorbox{orange}{Rep, Nation}, \\ \colorbox{yellow}{country, county, government, civil}\end{tabular}   & \begin{tabular}[c]{@{}l@{}}\colorbox{pink}{boys, editor, Dr, Jack, men, Judge, him}, \\ work, Chair, politics, religion, justice, \\ brave, \colorbox{pink}{Scott, }\colorbox{orange}{business, manager},\\ \colorbox{orange}{secret, ary,}\colorbox{yellow}{ she, mother}, \\ love, house, hand, \colorbox{yellow}{dress, Virgin}\end{tabular} & \begin{tabular}[c]{@{}l@{}}\colorbox{pink}{six, four, two, hundred, fifty, }\\\colorbox{pink}{ thousand, many, few, several, }\\\colorbox{pink}{ every, another, anything, }\\ \colorbox{pink}{ enough majority, ton, }\colorbox{pink}{massive}\\ \colorbox{lime}{great, remarkable,}  \colorbox{yellow}{generally}\end{tabular}     \\ \bottomrule
\end{tabular}
\caption{Visualization of the memory updating. We take the AM linked to word embedding layer and log the memory assignment of each token during the CLM on PG-19. As discussed in \cref{sec:visulization}, the original concept (which is written into memory earlier, we show them in \colorbox{pink}{red}) can shift slightly to a closed new one (colored in \colorbox{yellow}{yellow}) during the consolidation, caused by transition words (colored in \colorbox{orange}{orange}) or polysemous words (colored in \colorbox{lime}{green}).}
\label{tab:slot_vis}
\end{table*}

Results shown in \cref{tab:ablation} reveal that \ours w/o Read performs significantly worse compared with full model, emphasizing the crucial role of Read function in ensuring semantic relevance. When a random cache is returned in this variant, it might provide limited or even harmful information for current modeling. \ours w/o Recency also shows a notable performance dip over the full \ours model, confirming the essential role of maintaining the proper recency in the memory.
Variations in token consolidation and replacement also impact performance, resulting in different performance drops compared to the full approach. A larger decrement can be expected if the memory size gets smaller or the modeling corpus gets longer. These findings suggest \ours's optimal performance relies on the combination of  relevance, recency, novelty, and effective consolidation.
Please refer to \cref{sec:more_ab} to see more discussions on ablation study.

\subsection{Visualization:  Content Stored in AM}
\label{sec:visulization}

This section visualizes the contents in the AM's memory banks during CLM, to provide insights into memory usage dynamics. \cref{tab:slot_vis} displays the updates of six slots by processing input tokens over time.


First,  we identified two key types of memory slots in \ours: 
1. \textit{Functional Slots} that capture lexical, syntactical, or grammatical aspects of tokens, as seen in the top rows of \cref{tab:slot_vis}, related to modeling language structures and rules; and 2. \textit{Semantic Slots} in the bottom rows which captures the semantic essence of inputs. Tokens are assigned to slots based on their functional or semantic relevance, aligning with previous findings that embeddings from different layers or attention heads in Transformer-based models can specialize in different language aspects \cite{vig2019analyzing}.  We also notice each slot has consistent modeling rules. For instance, despite the similar functional purposes, prefix and suffix tokens are allocated to separate slots. This indicates that \ours can detect similarities across various dimensions and the nuances within a single category.



Secondly, we notice each slot has consistent modeling rules. For instance, despite serving similar functional purposes to construct a word, prefix and suffix tokens are allocated to separate slots, as illustrated in \cref{tab:slot_vis}.  This demonstrates that \ours can detect similarities across various dimensions and within the nuances of a single category.

Lastly, we notice that slight concept shifts within slots occur during consolidation. Examples in the bottom rows of \cref{tab:slot_vis} shows 
 transitioning from \textit{federations} to \textit{civilizations} in slot $60$,  from \textit{masculine} to \textit{feminine} terms in Slot $394$, and from specific \textit{numbers} to \textit{quantitative adverbs} in Slot $7275$. 
These changes arise from context-dependent updates and the semantic diversity of words, where transitional tokens like ``\textit{business, manager, secret, ary}'' and polysemous words like ``\textit{great}'' influence the shifts in slot focus. We view this concept transitions as beneficial, as they facilitate an efficient consolidation while preserving recency. If these cumulative transitions lead to a significant change in the slot's mode, the slot can be replaced by the new token in the future rounds, as part of the novelty mechanism.

\subsection{\ours  Enhances Infrequent Word Modeling}
In this section, we answer the question that which words benefit from long-term knowledge in \ours. Following \cite{compressivetrans}, we categorize test tokens based on their frequency in the training set. The tokens are grouped into different frequency buckets, and we calculate the average perplexity for each group. 

Results are shown in \cref{tab:infrequent_ppl}. we notice that all tokens gain at least 19\% improvements over the base model. Among them, the high frequency tokens (frequency $>$ 10k), which constitute the majority of the test set, exhibit a 22.1\% improvement. The largest improvement (25.3\%) is observed in the group of rare tokens. This improvement suggests augmenting language models with mechanisms like \ours can be a viable approach to better address the challenges associated with rare token modeling.




\section{Conclusion}
We introduce \ours, a \textbf{C}onsolidated \textbf{A}ssociative \textbf{M}emory \textbf{E}nhanced \textbf{Lo}ng \textbf{T}ransformer, to handle long dependency modeling without the need for training. \ours has a model-agnostic design, allowing seamless integration into different language models. Experimental results prove its effectiveness, with the long-context language modeling perplexity significantly reduced (by up to 29.7\%), and superior performance is consistently obtained even with a tiny input window of 128 tokens. 
Future research directions connecting AM and LLMs involve improving the AM design (e.g., automatically learning a Write function) or tackling other long context modeling tasks (e.g., long document summarization or advanced reasoning).

\bibliography{example_paper}
\bibliographystyle{icml2022}

\newpage
\appendix
\onecolumn
\section{Appendix}

\subsection{Prompt Examples in In-context learning Task}
\label{Appendix:ICL_examples}

When constructing long prompt in ICL QA experiment on TREC, for each test question \texttt{Test\_Question}, we first randomly select $P$ data points from TREC training set, where each of them follows a style of: \texttt{Question: {Sentence}
Answer: {Label}},  

We then concatenate these $P$ examples, after which we append the question in the end, following the template:

\texttt{Question: {Sentence} Answer: {Label}; \\Question: {Sentence} Answer: {Label};\\ Question: {Sentence} Answer: {Label};\\ …\\ Question: {Sentence} Answer: {Label};\\ Question: Test\_Question Answer: <generation starts here>} 

We show a long test prompt example here:

\texttt{Question: How did serfdom develop in and then leave Russia ? Answer: DESC:manner; \\ Question: What was cash-conscious Colonel Edwin L. Drake the first to drill ? Answer: DESC:manner; \\ Question: What is `` dew point '' ? Answer: DESC:def; Question: Who invented Make-up ? Answer: HUM:ind; Test Question: What 's the only work by Michelangelo that bears his signature ? Answer: ENTY:cremat; Question: how much money does a back injury lawsuit get ? Answer: ENTY:cremat; Question: What is the busiest air travel season ? Answer: NUM:date; Question: Why do oceans contain salt water instead of fresh water ? Answer: DESC:reason; Question: How can you prevent it ? Answer: }

We input prompts into \ours for a continuous generation. The option that has the best perplexity among all options will be used as the prediction generated to the question.

\subsection{Experiment Details}
\label{Appendix: hyper}
\paragraph{Environments}
All transformers-based languague models are implemented based on the HuggingFace\footnote{\url{https://huggingface.co/models}} libraries (version 4.34.0) or the officially released Github Repos. All codes are implemented with Python 3.10.12 and PyTorch 2.2.0 with CUDA 12.1.0. We run experiments with 2 NVIDIA A100 GPUs, one for language model inference and one for hosting the memory banks. Each has memory of 80GB.
\paragraph{Hyper-parameters}
In CLM tasks, we set batch size to be 4. In ICL task, we conduct continuous generation for each test prompt one by one. For the similarity hyper-parameter $R$, we conduct a hyper-parameter study on wikitext-103 and use $R = 0.93$ for all experiments. We show the study in the next section. 
\subsection{More Ablation Studies}
\label{sec:more_ab}
\subsubsection{Ablation: Different Choices of $R$}
We conduct hyper-parameter study for similarity threshold $R$ on a subset of wikitext-103 validation set, in which the examples are randomly sampled. We take LLaMa2 models with input length to be 128. The results are shown in \cref{tab:R_hyper}. 

From the results, we notice the best $R$ is within $[0.9, 0.95)$. Therefore we use 0.93 in our experiments. 
\begin{table}[ht]
\centering
\begin{tabular}{@{}cc@{}}
\toprule
$R$     & Perplexity \\ \midrule
R=0.1 & PPL=18.72  \\
R=0.2 & PPL=18.71  \\
R=0.3 & PPL=18.71  \\
R=0.4 & PPL=18.69  \\
R=0.5 & PPL=18.67  \\
R=0.6 & PPL=18.46  \\
R=0.7 & PPL=17.35  \\
R=0.8 & PPL=15.30  \\
R=0.9 & PPL=\textbf{14.38 } \\
R=9.5 & PPL=14.90  \\ \bottomrule
\end{tabular}

\caption{Hyper-parameter study for $R$ on the validation subset of wikitext-103}
\label{tab:R_hyper}
\end{table}

\subsubsection{Ablation: Different Choices of Similarity Function in Read Operation}
We conduct an ablation study on the similarity function  in Read operation. Similarly, we randomly sampled a subset data from the validation set of wikitext-103. We conduct evaluation experiments with cosine similarity and euclidean similarity. We use input window  with 128 tokens. Note in this experiment we use LLaMa1-7B. The results are shown in \cref{tab:ab_similarity}. We notice cosine similarity gives the best performance and we use cosine similarity in our other experiments. 
\begin{table}[]
\centering
\begin{tabular}{@{}lc@{}}
\toprule
                                            & Perplexity     \\ \midrule
\ours + Cosine Similarity    & 16.96          \\
\ours + Euclidean Similarity & \textbf{17.45} \\ \bottomrule
\end{tabular}
\caption{Analysis of similarity function on a subset of wikitext-103 validation set. }
\label{tab:ab_similarity}
\end{table}


\subsubsection{Ablation: Different Memory Sizes}
\begin{table}[h]
\centering
\begin{tabular}{@{}cccc@{}}
\toprule
Model                      & Input Context & Memory Size & Perplexity \\ \midrule
\multirow{2}{*}{LLaMa2-7B} & 512           & None        & 9.84       \\
                           & 2048          & None        & 7.88       \\ \midrule
\multirow{4}{*}{CAMELoT}   & 512           & 4096        & 7.42       \\
                           & 2048          & 4096        & 7.22       \\
                           & 512           & 10k         & 7.24       \\
                           & 2048          & 10k         & \textbf{7.10 }      \\ \bottomrule
\end{tabular}
\caption{Language Modeling performance on PG19 with different sizes of memory banks and different input lengths.}
\label{tab:memory_different_size_length}
\end{table}

\label{Appendix:more_memory_size}

In this section, we analyze how the size of the memory affects \ours. We compare its performance on the PG-19 dataset using two configurations: one with 4,096 memory slots and another with 10k slots. The findings are presented in \cref{tab:memory_different_size_length}.


With each memory slot designed to hold a unique mode of information, increasing the number of slots allows \ours to capture a wider range of knowledge. As a result, the version with 10k slots outperforms, showing a notable improvement in test perplexity -- 26.4\% for inputs of 512 length and 9.9\% for 2048 length relative to the base model.

However, the 4,096-slot configuration also performs strongly, with only slightly lower improvements (24.6\% and 8.4\%, for the same input lengths) than \ours with 10k slots. This good performance demonstrates that the effectiveness of \ours does not solely rely on the quantity of data modes it can hold in its memory, but also on how it manages and utilizes this data through mechanisms like consolidation and novelty. This balance ensures \ours remains effective across various memory sizes and input lengths, maintaining stability and efficiency.

\end{document}